\documentclass[conference]{IEEEtran}
\usepackage{multirow}

\usepackage{graphicx}
\usepackage{amsmath}
\usepackage{amssymb}
\usepackage{multirow}
\usepackage{booktabs}
\usepackage{url}
\usepackage[table]{xcolor}
\usepackage{xcolor}
\usepackage{pgfplotstable}

\def\BibTeX{{\rm B\kern-.05em{\sc i\kern-.025em b}\kern-.08em
		T\kern-.1667em\lower.7ex\hbox{E}\kern-.125emX}}
\def\cca#1{%
	\pgfmathsetmacro\calc{100-(#1-0)*100/(1-0)}%
	\edef\clrmacro{\noexpand\cellcolor{green!\calc}}%
	\clrmacro%
	\ifdim \calc pt>50pt\color{black}\fi{#1}%
}

\def\ccag#1{%
	\pgfmathsetmacro\calc{25}%
	\edef\clrmacro{\noexpand\cellcolor{green!\calc}}%
	\clrmacro%
	\ifdim \calc pt>50pt\color{black}\fi{#1}%
}

\def\ccar#1{%
	\pgfmathsetmacro\calc{25}%
	\edef\clrmacro{\noexpand\cellcolor{red!\calc}}%
	\clrmacro%
	\ifdim \calc pt>50pt\color{black}\fi{#1}%
}
\begin{document}

\title{Decision Support for Video-based Detection of Flu Symptoms}
	\author{\IEEEauthorblockN{ 
			{Kenneth Lai$^{1}$} and Svetlana N. Yanushkevich$^{1}$
		}
		\IEEEauthorblockA{
			\textit{$^1$Biometric Technologies Laboratory, Department of Electrical and Computer Engineering,} \textit{ University of Calgary, Canada} \\
			Web: http://www.ucalgary.ca/btlab, E-mail: \{kelai,syanshk\}@ucalgary.ca}\newline
	}

\maketitle

\thispagestyle{empty}

\begin{abstract}
	The development of decision support systems is a growing domain that can be applied in the area of disease control and diagnostics.  Using video-based surveillance data, skeleton features are extracted to perform action recognition, specifically the detection and recognition of coughing and sneezing motions.   Providing evidence of flu-like symptoms, a decision support system based on causal networks is capable of providing the operator with vital information for decision-making.  A modified residual temporal convolutional network is proposed for action recognition using skeleton features.  This paper addresses the capability of using results from a machine-learning model as evidence for a cognitive decision support system.  We propose risk and trust measures as a metric to bridge between machine-learning and machine-reasoning.  We provide experiments on evaluating the performance of the proposed network and how these performance measures can be combined with risk to generate trust.
\end{abstract}


\section{Introduction}\label{sec:introduction}
\begin{figure}[!ht]
	\begin{center}
		\includegraphics[width=0.99\columnwidth]{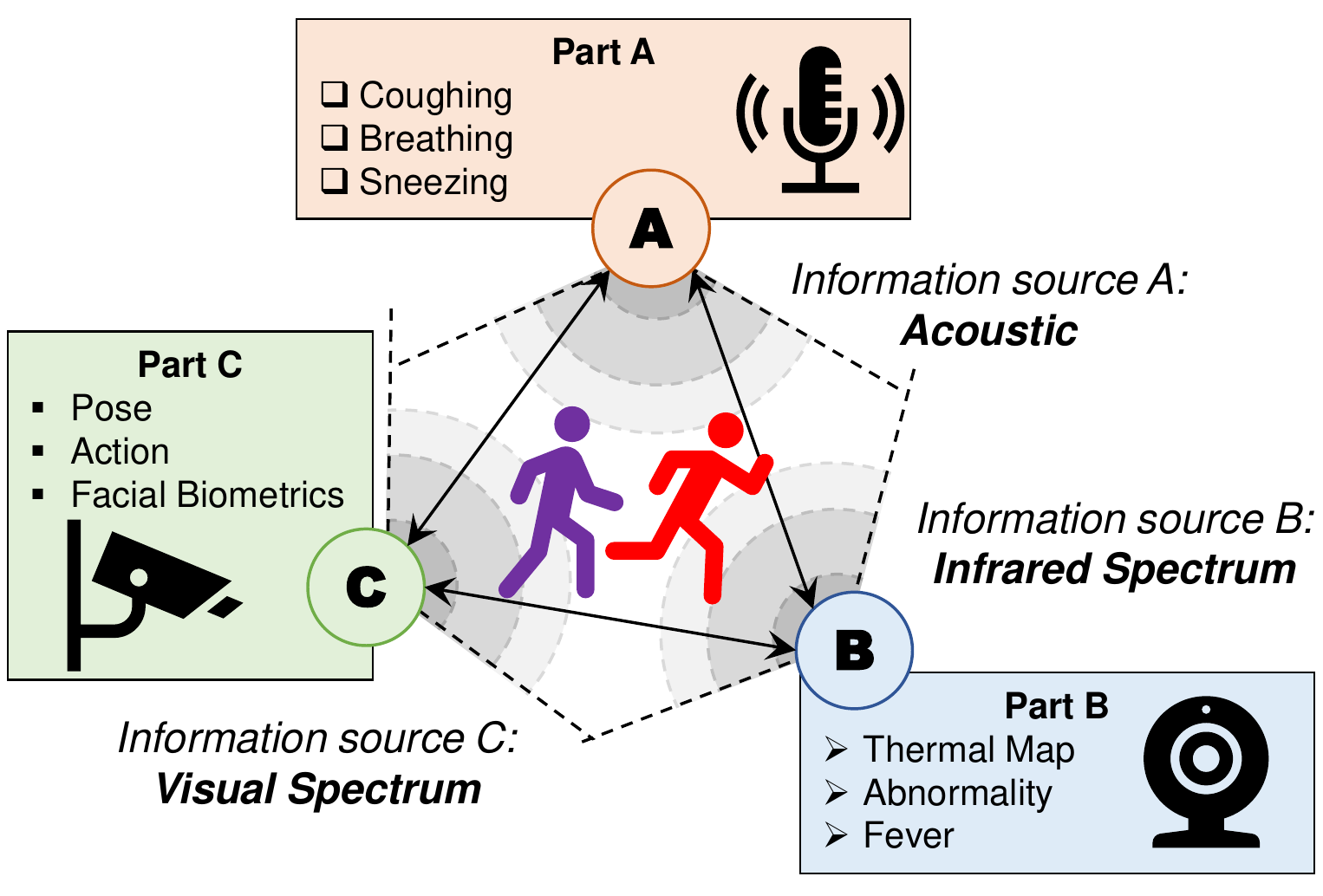}
	\end{center}
	\caption{An autonomous decision support system for human subject action detection and classification performs analysis and fusion of data from multiple sensors.}
	\label{fig:perception}
\end{figure}

Human body action recognition is one of the most active research topics in the area of pattern recognition.  Action recognition is required in application such as human behavior monitoring, human activity analytics, sport medicine and kinesiology. It is performed on  data from videos, images, or wearable sensors.  In this paper, we propose an approach to action recognition and its meta-analysis that provides output in the context of a decision support system (Figure \ref{fig:perception}). The meta-analysis data is understood here as risk, trust and reliability of the automated tool decision, as well as analysis of biases in those decision. This is required, in particular, to assist in disease  detection,  monitoring and predicting. This process is often a difficult task due to many variables such as time and uncertainty of data sources.  The proposed  decision support system combines machine learning and reasoning techniques to improve the performance of the system and supply  the risk associated with each prediction.

Recent research in the domain of action recognition is mainly based on videos, typically evaluated on various large video datasets such as the UCF101 \cite{soomro2017ucf101} and HMDB51 \cite{kuehne2011hmdb} database.  Action recognition is a difficult problem to solve due to the complexity of actions in videos. A long-term temporal convolutional neural network is proposed in \cite{varol2017long} to recognize actions reaching a performance of 92.7\% and 67.2\% for the UCF101 and HMDB51 database, respectively.  Similarly, \cite{feichtenhofer2017spatiotemporal} proposes a spatiotemporal multiplier network for video-based action recognition.  Other techniques such as combining convolutional neural networks with long-short-term-memory recurrent neural networks \cite{donahue2015long}, two streams convolutional neural networks \cite{simonyan2014two}, 3D convolutional neural networks \cite{tran2015learning}, and temporal segment networks \cite{wang2016temporal} have been proposed for action recognition on videos.

Databases such as UTD-MHAD \cite{chen2015utd} provides multi-modal information, including RGB videos, depth videos, skeleton points, and inertial signals.  Techniques that uses depth data is proposed in \cite{rahmani2016histogram} and \cite{rahmani20163d}.  Also, \cite{ahmad2019human} proposes to use CNN to recognition actions using depth sensor data. Similarly, a method proposed by \cite{chen2018distilling} combines knowledge distillation and smartphone data to recognition different actions. Other techniques for action recognition, specifically using skeleton features is proposed in \cite{rahmani2017learning} and \cite{ke2017new}.

The main objective of this paper is to implement a framework for action recognition, targeting specifically the classification of flu-like symptoms and how to incorporate the results of classification for a decision support system.  The proposed framework uses skeleton data to perform classification through the use of temporal convolution networks, residual connections, and knowledge distillation.  Knowledge distillation was previously used by other works \cite{hinton2015distilling} for distilling knowledge from an ensemble of models to a single model but has not been adapted into a boosting configuration with the aim to minimize losses and bias in the framework.  Boosting is an ensemble method that operates by training a series of weak classifiers into a strong classifier with an end goal of reducing overall bias.

The paper is structured as follows: a framework of the proposed method is given in Section \ref{sec:framework}, the design of experiments and experimental results are provided in Section \ref{sec:experiments}, and  Section \ref{sec:conclusions} concludes the paper.

\section{Framework}\label{sec:framework}
In this paper, we propose a combination of residual connections, temporal convolution network, and knowledge distillation.  A temporal convolution network that uses residual connections is proposed in \cite{kim2017interpretable} for 3D human action analysis. 

\subsection{Residual-Temporal Convolution Network}
\begin{figure*}[!ht]
	\begin{center}
		\includegraphics[width=0.98\textwidth]{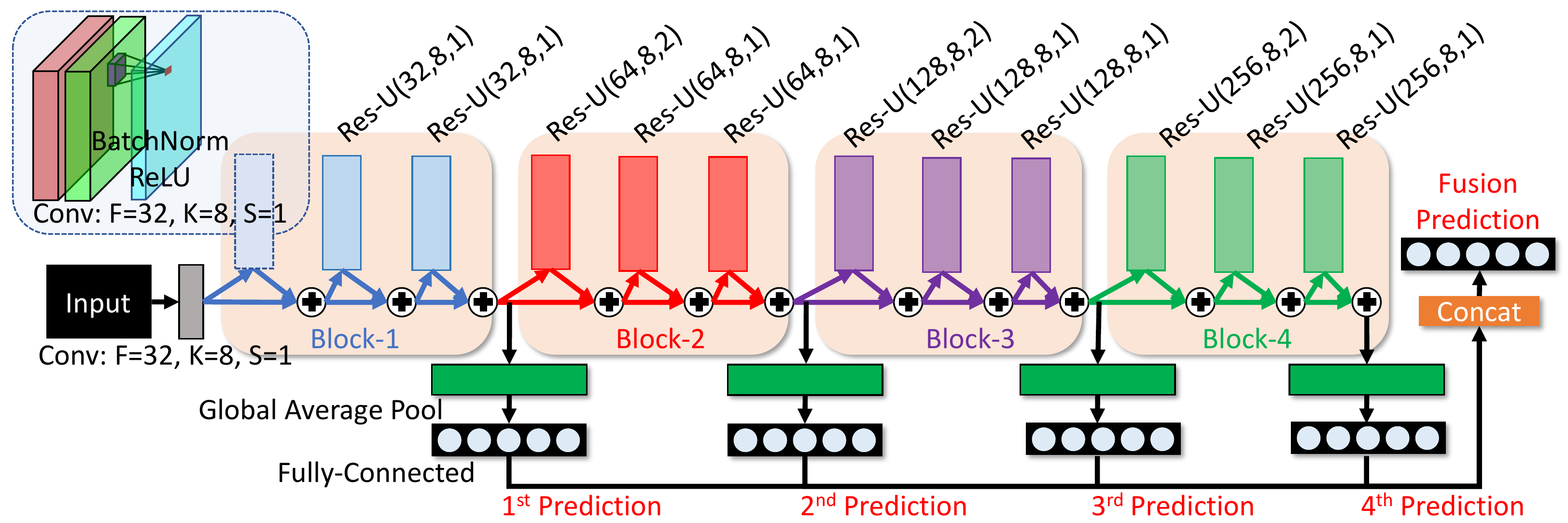}
	\end{center}
	\caption{The architecture of the Res-TCN.  The network contains four blocks of residual units. Each block produces a preliminary prediction that at the end contributes to a 4-block-fusion prediction. Each residual unit contains a series of three sub-block, where each sub-block is the combination of a BatchNorm, ReLu, and convolution layer.  A skip connection represented as an ``addition operator'' is introduced for each sub-block.}
	\label{fig:tcn}
\end{figure*}

Figure \ref{fig:tcn} illustrates the modified Residual-Temporal Convolution Network (Res-TCN) architecture for detecting falls (binary classification) and action recognition (multi-class classification).

The modified Res-TCN network is composed of four blocks of residual units.  Each residual unit is composed of three sets of sub-blocks where each sub-block is the combination of Batch Normalization (BatchNorm), Rectified Linear Unit (ReLu), and Convolution layers.  The sub-block structure is illustrated in Figure \ref{fig:tcn}.  $\texttt{Res-U}(32,8,1)$ represents a sub-block containing a convolutional layer with 32 filters ($F=32$), filter size of 8 ($K=8$), and stride of 1 ($S=1$).  Due to the residual connections, the output of every sub-block is summed up with the input to the sub-block.  Residual connections have shown to improve the interpretability of the results \cite{kim2017interpretable}.

\subsection{Knowledge distillation}

\emph{Knowledge distillation} is a term that describes  the process of transferring the ``dark'' knowledge from the one well-trained classifier to another ``weaker'' classifier.  ``Dark'' knowledge refers to the hidden information learned by the models, and can be revealed by calculating the softened probability based on a temperature $T$, as defined in equation below \cite{hinton2015distilling}: 
\begin{equation} \label{eq:sigma}
\sigma_{i,m}= \frac{\exp(Logit^i_m/T)}{\sum_{j}^{N}\exp(Logit^j_m/T)}
\end{equation}
where $N$ is the total number of classes, $T$ is the temperature parameter, $Logit^i_m$ is the $i^{th}$ class's \emph{Logit} (logarithm of the ratio of success to ratio of failure) output from $m$ network. When $T=1$, the resulting probability is the same as the result of a Softmax function. 

In this paper, we apply a knowledge distillation method called Fusion Knowledge Distillation (FKD) defined in \cite{kim2019feature}. FKD can be used to transfer the learned knowledge of the ensemble classifier to the individual block networks.  

\begin{figure}[!ht]
	\begin{center}
		\includegraphics[width=0.48\textwidth]{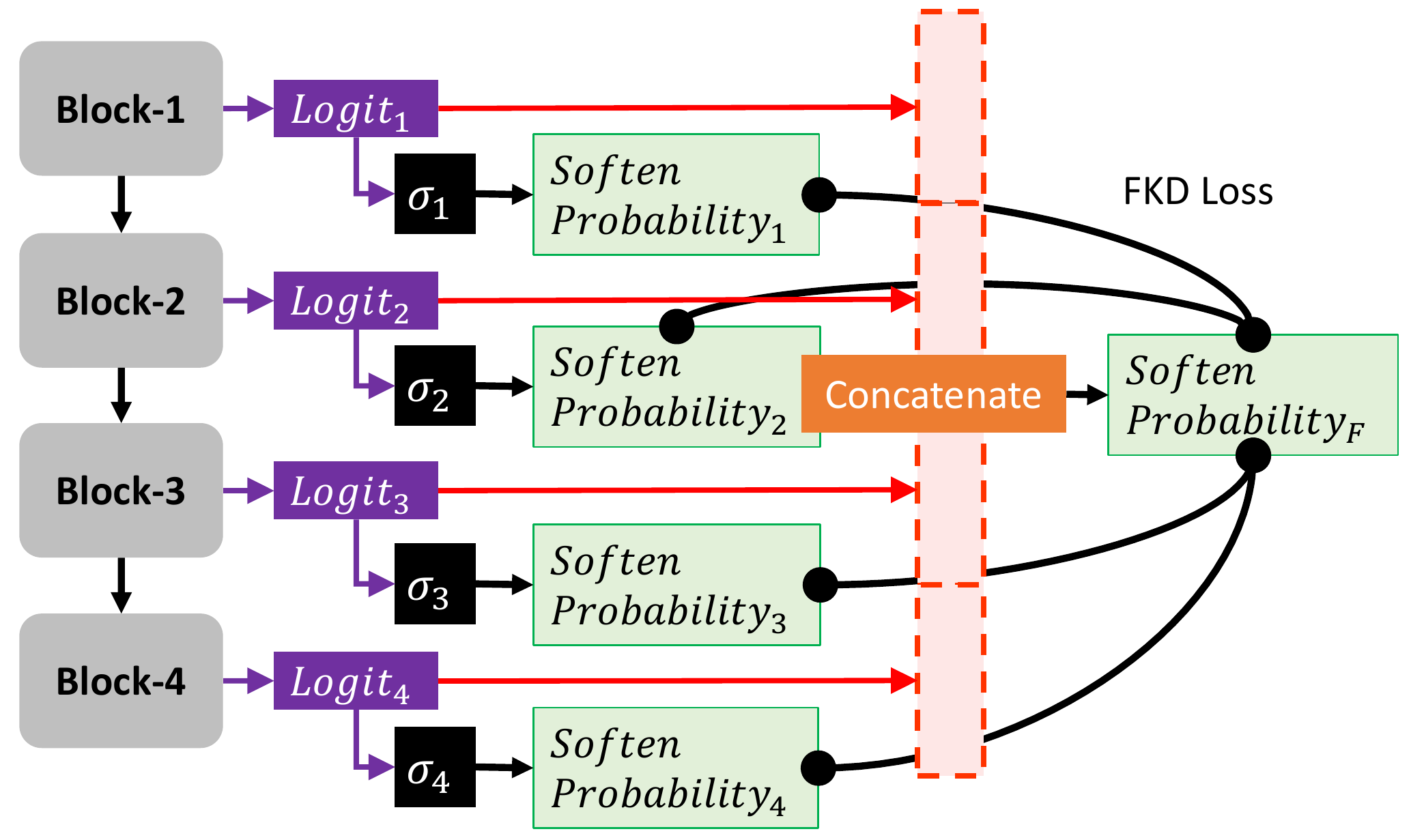} 
		\caption{The knowledge distillation between the fusion classifier and the block networks. The Logit from each block is obtained from the fully-connected layers.  The soften probabilities are calculated using Equation \ref{eq:sigma}, where $T=3$. FKD loss represents the Kullback-Leibler divergence between the softened probabilities from the block and fusion networks.  Equation \ref{eq:loss1} is used to calculate the FKD loss ($N=4$).}
		\label{fig:distillation}
	\end{center}
\end{figure}

Figure \ref{fig:distillation} illustrates the FKD loss that used to transfer knowledge between the classifiers.  FKD loss is defined as the Kullback-Leibler divergence between the softened distribution of the fusion classifier and the softened distribution of the individual networks.

The loss for the individual sub-network is the combination of the cross-entropy  loss and the FKD loss defined as follows:
\begin{equation} \label{eq:loss1}
\begin{split}
L_{m}= \underbrace{\sum_{i}^{N}\sigma_{i,f} \log(\frac{\sigma_{i,f}}{\sigma_{i,m}})}_\text{FKD Loss} \underbrace{- \sum_{i}^{N}y_i \log(\sigma_{i,m})}_\text{Cross-Entropy Loss}
\end{split}
\end{equation}
where $L_m$ is the loss for the $m$ block network, $L_f$ is the loss for the fusion classifier, $y_i$ is the truth label for the $i^{th}$ index in a one-hot-encoded label.

 An effective means to transfer the learned Logit from the fusion classifier to the block networks is by reducing the FKD loss between the block network and the fusion classifier. This transfer of knowledge allows for more optimal weight updates for the earlier stages of the network.

\subsection{Trust, Risk and Reliability}

In our study, we use the formalization of Trust as a function of risk of error in the autonomous system decision, as well  reliability of such decision:
\begin{equation}
\text{Trust} = \mathcal{F}(\text{Risk},\ \text{Reliability})
\end{equation}

Risk is estimated as a function $\mathcal{F}(\text{Impact}, \text{Probability})$ where \text{Probability} represents the error rates of the system, specifically the false non-match rate (FNMR) and the false match rate (FMR). The Risk in  this context is interpreted as a cost of decision error. In particular, the Risk of the  classification error is computed as follows:
\begin{eqnarray} \label{eq:risk}
\text{Risk}_{Error}&=& \underbrace{\text{Impact}_{\textit{FNMR}}}_\text{Cost of a FNMR} \times \underbrace{\text{Error}_{\textit{FNMR}}}_{\text{FNMR}} \nonumber\\
&+&\underbrace{\text{Impact}_{\textit{FMR}}}_\text{Cost of a FMR} \times \underbrace{\text{Error}_{\textit{FMR}}}_{\text{FMR}}\\
&=& \alpha \cdot \text{Error}_{\textit{FNMR}} + \beta \cdot \text{Error}_{\textit{FMR}}\nonumber
\end{eqnarray}
where $\alpha$ represents the \text{Impact} or cost of a false non-match and $\beta$ represents the \text{Impact} or cost of a false match in classification of human action, in our case.

In \cite{cohen1998trust}, the definition of Trust is dependent on both the qualitative and quantitative aspects of the system.  In this paper, we define Trust as the degree of confidence in a system decision, that is, the ability for the system to produce the correct prediction over many iterations. As such, we draw a parallel between the change in Trust and the change in  decision Reliability.

 In the context of classification, Reliability of the decision made by an autonomous system is defined {as the true match rate.  For rank-1  based classification (the decision is made upon the top found match), the true match rate is equivalent to the accuracy  of the system decision (Equation \ref{eq:acc}).}

In this paper, we focus on a change in decision Reliability, which is thought of as a bias:
\begin{eqnarray}\label{eq:reliability}
 \text{Bias}_{Reliability} &=&  \text{Reliability}_j - \text{Reliability}_i
\end{eqnarray}
where $\text{Reliability}_i$ represents the reliability of the system decision given condition $i$ and $\text{Reliability}_j$ represents the reliability of the system decision given condition $j$.  Given the rank-1 classification example considered in this paper, reliability of the system decision is computed as the accuracy of the decision.  Note that $\text{Bias}_{Reliability}$ can be positive which indicates an increase in reliability, or negative which means a decrease in reliability.

Similarly,  $\text{Bias}_{Risk}$ is defined as follows: 
\begin{eqnarray}\label{eq:trust}
\text{Bias}_{Risk} &=&  \text{Risk}_{i} - \text{Risk}_{j},
\end{eqnarray}
Note that $\text{Bias}_{Risk}$ as either a positive or a negative value representing the direction  of change in Risk when comparing decision error under conditions $i$ and $j$, respectively.

In this paper, we focus on capturing the change in the degree of Trust as it is a critical task in real-world applications.  As described in \cite{eastwood2018technology}, the performance in a simulated environment is wildly different from real-world applications.  It should be noted that  the reliability matrices can be constructed in order to help identify the biases within the system and instead of constructing the Reliability matrices directly, we provide the accuracy measures in Table \ref{tab:gp_performance} and \ref{tab:v_performance}. The construction and analysis of the Reliability matrices is out of scope of this paper. 

\subsection{Capturing a bias ensemble }

The existence of bias can deeply alter the performance of any classification algorithms. One common bias is attributed to the demographic of the dataset. An example is a dataset containing an imbalanced number of subjects for gender.  Given the list of the different cohorts (poses) and the overall performance of a dataset, we can divide the overall performance into separate performance measures categorized based on the individual cohorts. 

Analysis of the performance of the proposed action (cough and sneeze) classifier suggests that the experimental data or setup contains a bias ensemble.  The performance of the classifier can be greatly improved by addressing such biases. We applied machine reasoning to analyze the results of the proposed machine learning approach for classifying human actions. Machine reasoning is a probabilistic reasoning technique that is based on causal graph structures called causal networks.

\begin{figure}[!ht]
	\begin{center}
		\includegraphics[width=0.35\textwidth]{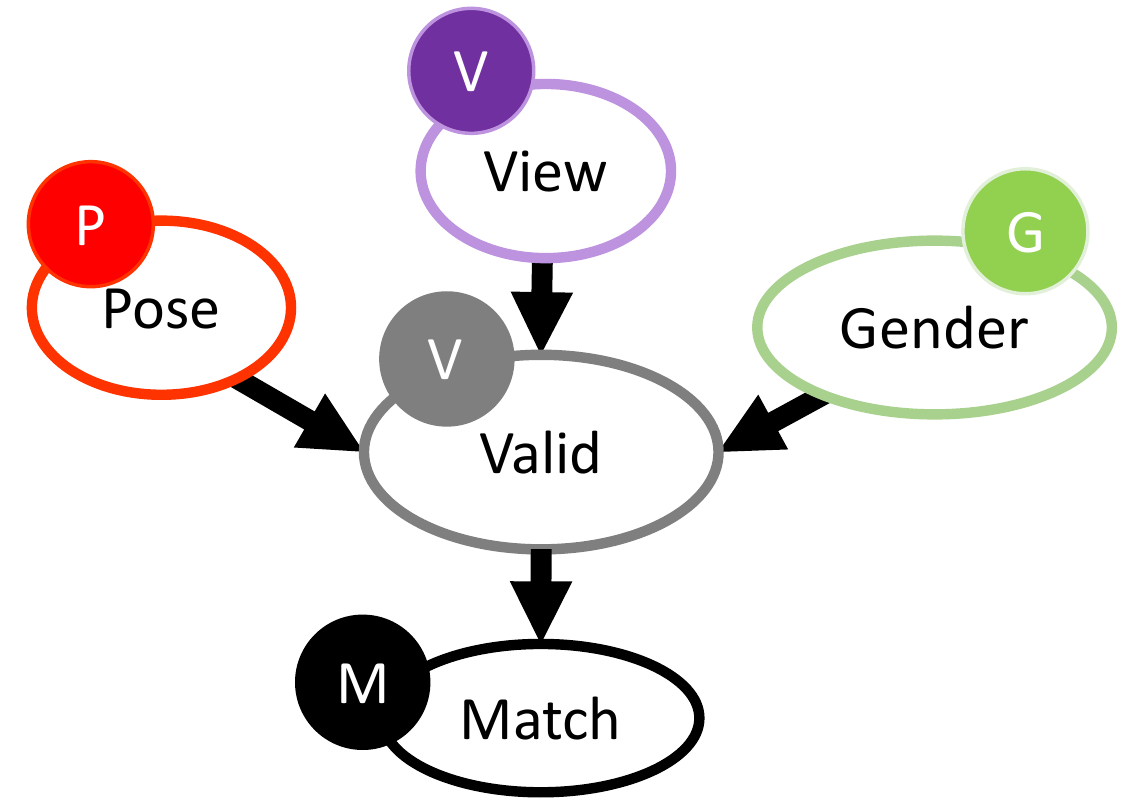}
		\caption{Causal Network of an ensemble bias in action classification.}
		\label{fig:Biases}
	\end{center}
\end{figure}

A causal network in Figure \ref{fig:Biases} captures the biases that are believed to influence the performance of flu-symptom detection/classification. These biases include human subject attributes such as view $V$, gender $G$, and pose $P$.  The parent nodes to the ``Valid'' node represent the different bias attributes that affect the recognition performance.  The ``Valid'' node represents the probability of the classifier in predicting a positive (valid) or negative (invalid) action. The ``Match'' node determines whether the positive or negative prediction matches the ground truth label.

Given the causal network and the corresponding Conditional Probability Tables (CPTs), a Bayesian network is built. Posterior probabilities can be calculated by applying Bayesian inference using the Bayesian network, prior probabilities, and the current observation. This is the mechanism proposed in this paper to analyze the influence of bias attributes on the performance of the flu detection system.

\section{Experimental results and discussions}\label{sec:experiments}

We conducted an experimental study aiming at 1) recognizing and 2) assessing biases in the task of human action classification.

The Res-TCN model to classify actions is trained using Adam CNN with cosine annealing for one cycle, the base learning rate $lr=0.001$ trained for 1000 epoch.  Adam optimizer with default parameters o $\beta_1=0.9$, $\beta_2=0.999$, and $\epsilon=1e^{-08}$ is used to train the network for the various validation methods.

The performance of each model is evaluated using the following measures:
 \textit{Accuracy} (Acc.),  \textit{Precision} (Prec.),  \textit{Sensitivity} (Sens.), and \textit{Specificity} (Spec.) metric. They are defined below:

\begin{small}
\begin{eqnarray}\label{eq:acc}
\textit{Accuracy} &=& \frac{TP+TN}{FP+FN+TP+TN}\\
\textit{Precision}&=&\frac{TP}{TP+FP}\label{eq:pr}\\
\textit{Sensitivity} &=&\frac{TP}{FN+TP}\label{eq:se}\\
\textit{Specificity} &=&\frac{TN}{FP+TN}\label{eq:sp}
\end{eqnarray}
\end{small}

where $TP$ are true positives (correct prediction of action), $TN$ are true negatives (correct prediction of non-action), $FP$ are false positives (incorrect prediction of action), $FN$ are false negatives (incorrect prediction of non-action).

\subsection{Dataset}
We used the BII Sneeze-Cough Human Action Video Dataset (BIISC) \cite{thi2014recognizing} that consists of 20 different subjects performing various actions, including coughing and sneezing.  Each subject performed 8 actions, of which each one is performed from three different perspectives and at different poses.  Video for each action is collected using a camera at 10 frames per second at a resolution of 480x290 pixels.  All videos are pre-processed using HR-NET \cite{cheng2019bottom} for pose estimation to extract 17 skeleton point features.

\begin{figure}[!ht]
	\begin{center}
		\begin{tabular}{c}
			\includegraphics[width=0.45\textwidth]{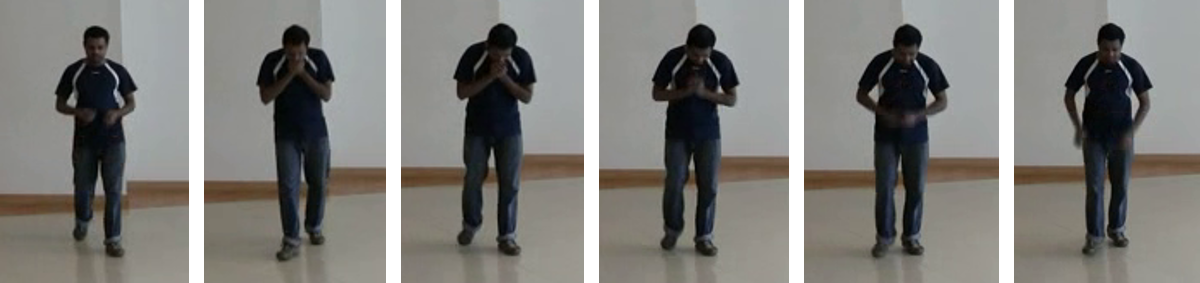} \\(a) \\
			\includegraphics[width=0.45\textwidth]{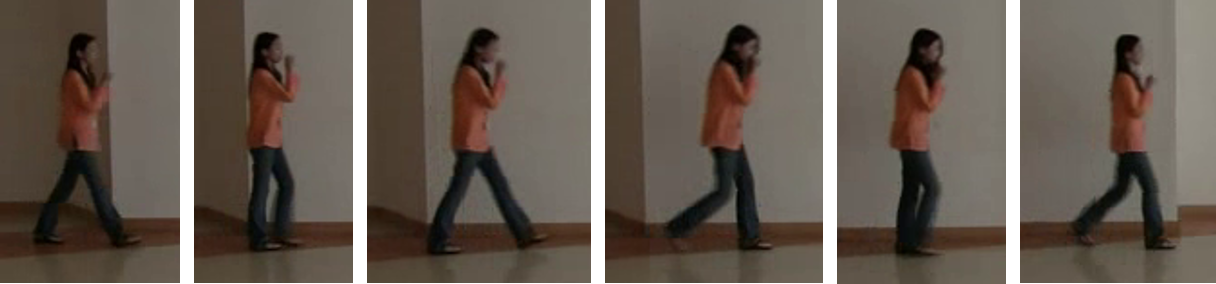} \\(b) \\
			\includegraphics[width=0.45\textwidth]{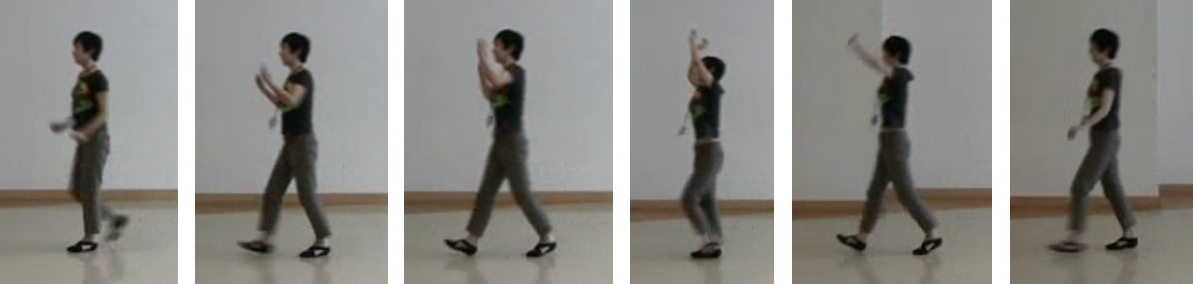} \\(c) \\
		\end{tabular}
		\caption{Sample frames of different actions from the BIISC dataset \cite{thi2014recognizing}: (a) coughing, (b) sneezing, and (d) stretching.}
		\label{fig:img}
	\end{center}
\end{figure}

The BIISC dataset was evaluated using the multiple methods include the method described in \cite{thi2014recognizing}. In \cite{thi2014recognizing}, the testing/training procedure is based on a pre-defined training and testing set. Specifically, subjects 2-6 are used for testing while the remaining subjects are used for training. In this paper, we aim to closely examine the influence of dataset bias, therefore we specifically separate the training and testing set based on selected attributes such as gender and pose.

\subsection{Action-Flu Classification}
The result of multi-class action classification is reported in Table \ref{tab:performance}.  The performance of the system is measured in terms of \textit{Accuracy} (Acc.),  \textit{Precision} (Prec.),  \textit{Sensitivity} (Sens.), and \textit{Specificity} (Spec.).  Each row in the table indicates a particular method used for performance evaluation.  In this paper, we proposed a boosted configuration that yields outputs after each residual block, resulting in 4 block outputs and a combined fusion result.  For each method, the performance is calculated based on the pre-defined training and testing set, specifically 5 subjects are used for testing while 15 subjects are used for training.

\begin{table}[!htb]
	\centering
	\caption{Multi-class Classification Performance}\label{tab:performance}
	\begin{tabular}{@{}c|cccc@{}}										
Method		&	Acc.	&	Sens.	&	Spec.	 & Prec.\\
		\hline	
Proposed (Block-1)	 & 	 0.767 	 & 	 0.779 	 & 	 0.843 	 & 	 0.767 	\\
Proposed (Block-2)	 & 	 0.817 	 & 	 0.813 	 & 	 0.851 	 & 	 0.817 	\\
Proposed (Block-3)	 & 	 0.817 	 & 	 0.827 	 & 	 0.851 	 & 	 0.817 	\\
Proposed (Block-4)	 & 	 0.810 	 & 	 0.826 	 & 	 \textbf{0.853} 	 & 	 0.810 	\\
Proposed (Fusion)	 & 	 \textbf{0.825} 	 & 	 \textbf{0.837} 	 & 	 0.852 	 & 	 \textbf{0.825} 	\\
		\hline
		HOGHOF+AMKII \cite{thi2014recognizing}&	 0.444 & 	0.644	 & 	- & 0.589	 \\ 
		cuboid+AMKII \cite{thi2014recognizing}&	 0.413 & 	0.621	 & 	- & 0.553	 \\ 
	\end{tabular}
\end{table}

Table \ref{tab:performance} indicates that the proposed method is significantly better than traditional feature extraction methods such as histogram of oriented gradient (HOG) and histogram of optical flow (HOF). The fusion result provides the highest accuracy (82.5\%), sensitivity (83.7\%), and precision (82.5\%).  The block-4 results yield the highest specificity (85.3\%).

\begin{figure}[!ht]
	\begin{center}
		\includegraphics[width=0.45\textwidth]{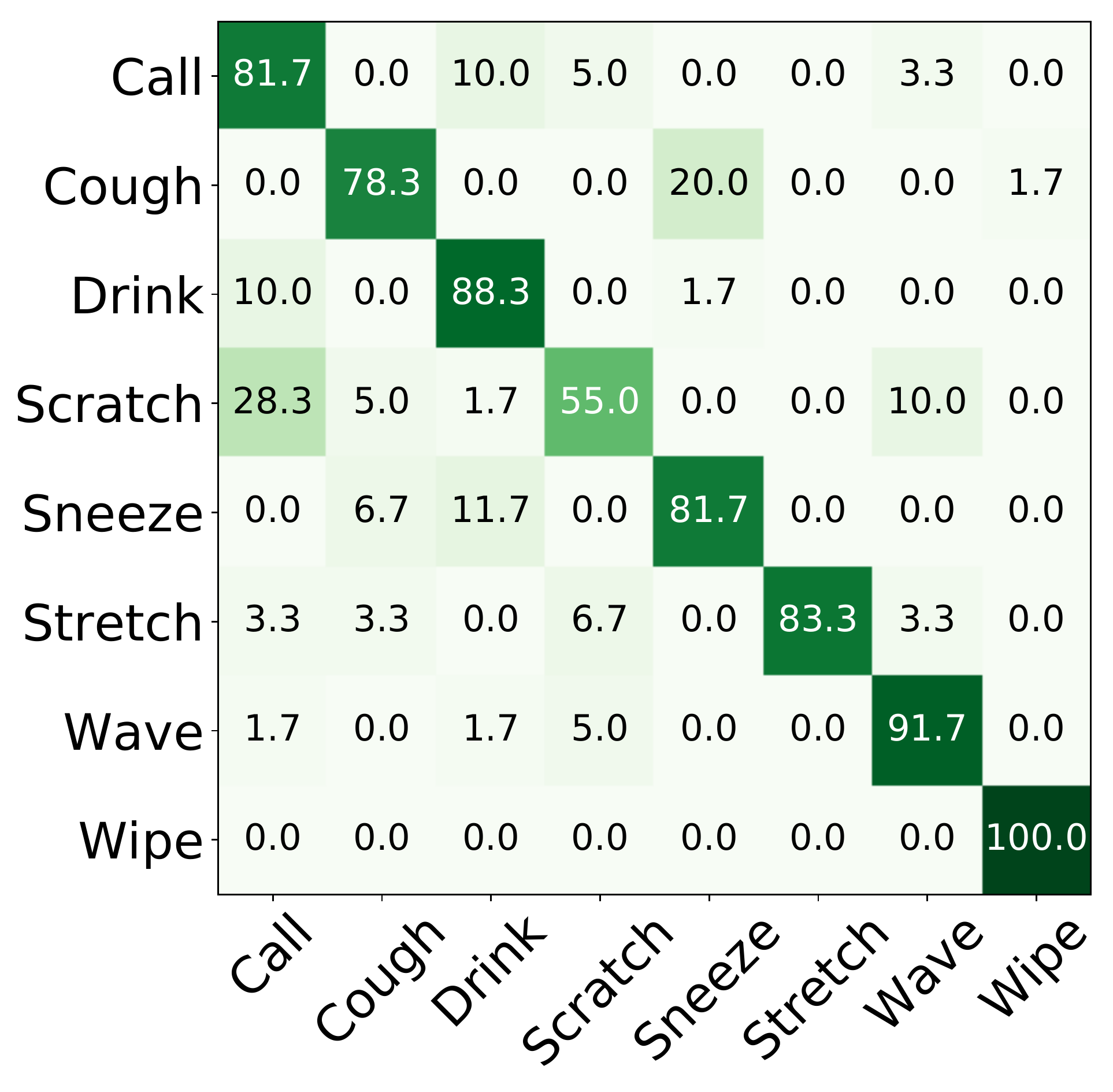}
	\end{center}
	\caption{A confusion matrix for multi-class action classification applied onto the BIISC dataset. There are a total of 8 actions performed by 20 different subjects.  The columns represent the final prediction of the model determined by the fusion result, while the rows represent the ground-truth action. The sum of each row is 100\% as it represents the total number of samples for that class.}
	\label{fig:cm}
\end{figure}

Figure \ref{fig:cm} illustrates a multi-class action recognition confusion matrix for the BIISC dataset.  Together, the columns (predicted result) and rows (ground-truth) represent the performance of a specific action given the model prediction and ground truth. The accuracy of the system can be computed using the diagonal values of the confusion matrix. Because in this dataset the number of samples for each class is the same, the arithmetic mean of the diagonal values also represents the balanced accuracy.  

In this section, we demonstrate that our proposed modified Res-TCN achieves significantly better performance when compared to traditional feature extraction methods.

\subsection{Risk and Biases}
Given the attributes within a dataset, a causal network (Figure \ref{fig:Biases}) can be created to calculate the risk associated with the  prediction provided by the model, and perform an inference to estimate the posterior probability associated with the prediction.  Using the prior probabilities given in Table \ref{tab:prior} and the causal relationship described in Figure \ref{fig:Biases}, we can estimate the influence of different attributes on the performance of the action classification system.  When this influence is combined with the error rates of the system, the risk resulting from these biases can be computed using Equation \ref{eq:risk}.

\begin{table}[!ht]
	\begin{center} 
		\caption{The prior probabilities corresponding to Figure \ref{fig:Biases}.}
		\begin{tabular}{cc}
			\begin{tabular}{c|cc}
				& \multicolumn{2}{c}{Gender} \\
				$G$ & Male & Female \\
				\hline
				\(\Pr(G)\) & 0.60 & 0.40 \\
			\end{tabular} &

			\begin{tabular}{c|cc}
				& \multicolumn{2}{c}{Pose} \\
				$P$ & Standing & Walking\\
				\hline
				\(\Pr(P)\) & 0.50 & 0.50	\\
			\end{tabular} \\
	& 	\\
			\multicolumn{2}{c}{\begin{tabular}{c|ccc}
			& \multicolumn{3}{c}{View} \\
			$V$ & Left & Center & Right \\
			\hline
			\(\Pr(V)\) & 0.33 & 0.33 & 0.33\\
		\end{tabular}} \\
			\label{tab:prior}
		\end{tabular} 
	\end{center}
\end{table}

Table \ref{tab:recognition} reports the performance of multi-class action recognition on the BIISC dataset partitioned into cohorts based on the different attributes (pose, view, and gender) in the dataset.  Performance is measured in terms of \textit{Accuracy} (Acc.),  \textit{Precision} (Prec.),  \textit{Sensitivity} (Sens.), and \textit{Specificity} (Spec.) and is evaluated based on the 5/15 subject split specified in \cite{thi2014recognizing}.

The overall performance of the model is given in the first row in Table \ref{tab:recognition}. Note that the performance for the female attribute is excluded as there were no female subjects in the test set.  Each attribute may represent a bias in the system as shown in Table \ref{tab:recognition}.  Given the baseline performance, positive biases are highlighted in blue, while negative biases are highlighted in red.  The performance of the different attribute/bias is computed based on the causal network (Figure \ref{fig:Biases}.

For example, significant bias is observed in the `view' attribute, the accuracy for the `left' view is significantly higher than the `center' view even though the prior probabilities (Table \ref{tab:prior}) for each view are the same.  This behavior illustrates a bias in the system, specifically the `left' views are easier to recognize as oppose to the `center' views.  A difference in 5.6\% accuracy is observed between the `left' and the `center' views.

\begin{table}[!htb]
	\centering
	\caption{Performance of Multi-class Classification for Various Attributes }\label{tab:recognition}
	\begin{tabular}{@{}cc|cccc@{}}
		\multicolumn{2}{c|}{Attribute}		&	Acc.	&	Sens.	&	Spec.	 & Prec.\\
		\hline
		\multicolumn{2}{c|}{Baseline (Fusion)}	 & 	 0.825 	 & 	 0.837 	 & 	 0.852 	 & 	 0.825	\\
		\hline							
\multirow{2}{*}{Gender}&Male 	& 	 0.825 	 & 	 0.837 	 & 	 0.852 	 & 	 0.825 	\\
&Female	&	-		&	-	&	-	&	-	\\
\hline
\multirow{2}{*}{Pose} &Stand& 	 0.825 	 & 	 \textcolor{blue}{0.848} 	 & 	 0.852 	 & 	 0.825 	\\
&Walk& 	 0.825 	 & 	 \textcolor{red}{0.825} 	 & 	 0.852 	 & 	 0.825 	\\
\hline
\multirow{3}{*}{View}&Center& 	 \textcolor{red}{0.794} 	 & 	 \textcolor{red}{0.796} 	 & 	 \textcolor{red}{0.847} 	 & 	 \textcolor{red}{0.794} 	\\
&Left& 	 \textcolor{blue}{0.850} 	 & 	 \textcolor{blue}{0.876} 	 & 	 \textcolor{blue}{0.856} 	 & 	 \textcolor{blue}{0.850} 	\\
&Right& 	 \textcolor{blue}{0.831} 	 & 	 \textcolor{blue}{0.850} 	 & 	 \textcolor{blue}{0.853} 	 & 	 \textcolor{blue}{0.831} 	\\			
	\end{tabular}
\end{table}

After capturing the different biases in the system, it is necessary to compute the risk associated with each bias.  Given Equation \ref{eq:risk} and balanced cost ($\alpha=1$ and $\beta=1$), we compute the baseline risk of the classification error as follows:
\begin{eqnarray}
	\text{Risk}_{{Error}} &=& \alpha \cdot \overbrace{\text{Error}_{\textit{FNMR}}}^{\text{1-{Sensitivity}}}+ \beta \cdot \overbrace{\text{Error}_{\textit{FMR}}}^{\text{1-{Specificity}}}\nonumber\\
	&=&1-0.837+1-0.852=\fbox{0.311}\nonumber
\end{eqnarray}

Note that an error resulting from FNMR and FMR can be computed based on the sensitivity and specificity of the system. The risk associated with the `left' pose is \fbox{0.268}, while the risk for the `center' pose is \fbox{0.357}. Computing the difference between baseline performance with the two poses illustrates the idea of positive and negative bias.
\begin{eqnarray}
	\text{Bias}_{Risk}({\text{Base}\to\text{Left}})&=&\text{Risk}_{\textit{Baseline}}-\text{Risk}_{\textit{Left view}}\nonumber\\
	&=&0.311-0.268 = \textcolor{blue}{\fbox{0.043}}\nonumber\\
		\text{Bias}_{Risk}({\text{Base}\to\text{Center}})&=&\text{Risk}_{\textit{Baseline}}-\text{Risk}_{\textit{Center view}}\nonumber\\
	&=&0.311-0.357 = \textcolor{red}{\fbox{-0.046}}\nonumber
\end{eqnarray}

\begin{table*}[!ht]
	\centering
	\caption{Multi-Class Action Recognition Performance Evaluated based on Gender and Pose Testing}
	\label{tab:gp_performance}
	\begin{tabular}{c||cccc|cccc||cccc|cccc}
		&	\multicolumn{4}{c|}{Male} & \multicolumn{4}{c||}{Female} &	\multicolumn{4}{c|}{Walking} & \multicolumn{4}{c}{Standing}\\
		
		&	Acc.	&	Sens.	&	Spec.	&	Prec.	&	Acc.	&	Sens.	&	Spec.	&	Prec.&	Acc.	&	Sens.	&	Spec.	&	Prec.	&	Acc.	&	Sens.	&	Spec.	&	Prec.\\
		\hline
		Block-1	&	 0.635 	 & 	 0.631 	 & 	 0.816 	 & 	 0.635 	 & 	 0.703 	 & 	 0.751 	 & 	 0.831 	 & 	 0.703 	 & 	 0.716 	 & 	 0.737 	 & 	 0.833 	 & 	 0.716 	 & 	 0.746 	 & 	 0.738 	 & 	 0.839 	 & 	 0.746 	 \\ 
		Block-2	&	 0.674 	 & 	 0.674 	 & 	 0.825 	 & 	 0.674 	 & 	 0.743 	 & 	 0.770 	 & 	 0.839 	 & 	 0.743 	 & 	 \textbf{0.796} 	 & 	 0.801 	 & 	 \textbf{0.848} 	 & 	 \textbf{0.796} 	 & 	 0.813 	 & 	 0.811 	 & 	 0.850 	 & 	 0.813 	 \\ 
		Block-3	&	 0.692 	 & 	 0.686 	 & 	\textbf{0.829} 	 & 	 0.692 	 & 	 0.757 	 & 	 0.760 	 & 	 0.841 	 & 	 0.757 	 & 	 0.794 	 & 	 0.803 	 & 	 0.847 	 & 	 0.794 	 & 	 0.814 	 & 	 \textbf{0.813} 	 & 	 \textbf{0.851} 	 & 	 0.814 	 \\ 
		Block-4	&	 \textbf{0.695} 	 & 	 \textbf{0.695} 	 & 	 0.828 	 & 	 \textbf{0.695} 	 & 	 0.757 	 & 	 0.750 	 & 	 0.842 	 & 	 0.757 	 & 	 0.776 	 & 	 0.791 	 & 	 0.844 	 & 	 0.776 	 & 	 0.803 	 & 	 0.804 	 & 	 0.849 	 & 	 0.803 	 \\ 
		Fusion	&	 0.690 	 & 	 0.687 	 & 	 0.828 	 & 	 0.690 	 & 	\textbf{0.760} 	 & 	 \textbf{0.775} 	 & 	\textbf{0.842} 	 & 	 \textbf{0.760}	 & 	 0.790 	 & 	 \textbf{0.805} 	 & 	 0.847 	 & 	 0.790 	 & 	 \textbf{0.815} 	 & 	 0.812 	 & 	 \textbf{0.851} 	 & 	 \textbf{0.815} 	 \\ 
	\end{tabular}
\end{table*}
\begin{table*}[!ht]
	\begin{center} 
		\caption{Multi-Class Action Recognition Performance Evaluated based on View Testing}
		\label{tab:v_performance}
		\begin{tabular}{c|cccc|cccc|cccc}
			&	\multicolumn{4}{c|}{Left} & \multicolumn{4}{c|}{Center} & \multicolumn{4}{c}{Right} \\
			&	Acc.	&	Sens.	&	Spec.	&	Prec.	&	Acc.	&	Sens.	&	Spec.	&	Prec.	&	Acc.	&	Sens.	&	Spec.	&	Prec.\\
			\hline
			Block-1	 & 	 0.755 	 & 	 0.741 	 & 	 0.841 	 & 	 0.755 	 & 	 0.783 	 & 	 0.778 	 & 	 0.846 	 & 	 0.783 	 & 	 0.806 	 & 	 0.816 	 & 	 0.849 	 & 	 0.806 	 \\ 
			Block-2	 & 	 0.806 	 & 	 0.797 	 & 	 0.849 	 & 	 0.806 	 & 	 0.813 	 & 	 0.802 	 & 	 0.850 	 & 	 0.813 	 & 	 0.858 	 & 	 0.861 	 & 	 0.857 	 & 	 0.858 	 \\ 
			Block-3	 & 	 \textbf{0.823} 	 & 	 0.817 	 & 	 0.852 	 & 	 \textbf{0.823} 	 & 	 0.817 	 & 	 0.811 	 & 	 0.851 	 & 	 0.817 	 & 	 0.877 	 & 	 0.877 	 & 	 \textbf{0.860} 	 & 	 0.877 	 \\ 
			Block-4	 & 	 0.805 	 & 	 \textbf{0.820} 	 & 	 \textbf{0.868} 	 & 	 0.794 	 & 	 0.811 	 & 	 0.805 	 & 	 0.850 	 & 	 0.811 	 & 	 0.873 	 & 	 0.874 	 & 	 0.859 	 & 	 0.873 	 \\ 
			Fusion	 & 	 0.822 	 & 	 0.812 	 & 	 0.852 	 & 	 0.822 	 & 	 \textbf{0.822} 	 & 	 \textbf{0.814} 	 & 	 \textbf{0.852} 	 & 	 \textbf{0.822} 	 & 	 \textbf{0.881} 	 & 	 \textbf{0.878} 	 & 	 \textbf{0.860} 	 & 	 \textbf{0.881} 	 \\ 
		\end{tabular} 
	\end{center}
\end{table*}
In this section, we capture the existence of bias within the different attributes in the BIISC dataset.  There is a significant bias for the view attribute, specifically, the `center' view offers much poorer results when compared to `left' or `right' views.  Given this bias, we can exploit this behavior by using only `left' and `right' view when operating the system.  In addition, we show how these biases can be combined to compute risk values.  The risk values provide the user with a quantitative measure to assess the performance of the system relative to the cost and impact, as opposed to performance metrics such as accuracy.  A significant negative bias can deter an operator even if the accuracy is very high, because the cost of error is too significant.

\subsection{Reliability and Trust}
The Reliability of the system decision depends on how often the system is able to correctly classify each action.  Given the rank-1 performance, the decision Reliability  is equivalent to the accuracy of the system.  The systems performance is usually evaluated  using some pre-defined testing and training sets \cite{thi2014recognizing}. However, in the previous section, we demonstrated the existence of bias in the dataset, and  that the pre-defined sets may not be optimal.  In this section, provide the results of cross-examining the performance of the system using various training and test sets.

Table \ref{tab:gp_performance} reports the performance of multi-class action recognition by controlling the gender and pose attributes.  Table \ref{tab:gp_performance} is separated into four panels: `Male', `Female', `Walking', and `Standing'.  The first panel in Table \ref{tab:gp_performance} shows the performance of the system when `Female' subjects are used for training while `Male' subjects are used for testing.  The second panel represents the performance of the system when `Male' subjects are used for training, while `Female' subjects are used for testing.  The third and fourth panel represents the procedure where `Walking' and `Standing' poses are used for testing, respectively.  Each row in the table indicates the output used for performance evaluation. 

The key finding from Table \ref{tab:gp_performance} is that the performance for each testing method, based on gender and pose, is worst than pre-defining subjects for testing.  This indicates that the testing protocol based on defining subjects for testing is insufficient in analyzing the general performance of the system.  Due to the inconsistency of performance between the testing protocols, the Reliability of the system decision should be much lower than expected.  

Table \ref{tab:v_performance} offers another testing protocol based on the `view' attribute.  A significant finding is that when testing with the `Right' view, the performance is exceptionally good.  This indicates that when subdividing the dataset based on the view attribute, using `left' and `center' views for training and `right' view for testing yields the most optimal result possible for this given system.  Due to the increase in performance when operating on `right' views, the overall reliability of the system decision is expected to be higher.

The overall trust in the autonomous system decision is depending on many factors such as decision Reliability and Risk of decision error.  In particular, $\text{Bias}_{Reliability}$ for `right' view can be computed based on Equation \ref{eq:reliability}:
\begin{eqnarray}
\text{Bias}_{Reliability} &=& \text{Reliability}_{{Right}}-\text{Reliability}_{{Base}}\nonumber\\
&=&0.881-0.825=\fbox{0.056} \nonumber
\end{eqnarray}
The $\text{Bias}_{Reliability}$ associated with the `right' view is positive, representing an overall gain in trust. This indicates that operating on classifying actions from the `right' view results in a more reliable system decision.  Other testing protocols using gender and pose testing results show the overall decrease in the decision Reliability, as the accuracy is lower.  

\section{Decision Support System}
Consider  an epidemic situation with flu-like symptoms. A decision support system provides the operator with a decision using collected evidence which can be video data, audio input and/or wearabale sensory data.  In typical scenarios, the evidence is collected by experts and, therefore, have lower error rates.  Bayesian inference can be used to calculate the chance of detecting a disease  such as flu given the symptoms/evidence.

\begin{figure}[!ht]
	\begin{center}
		\includegraphics[width=0.45\textwidth]{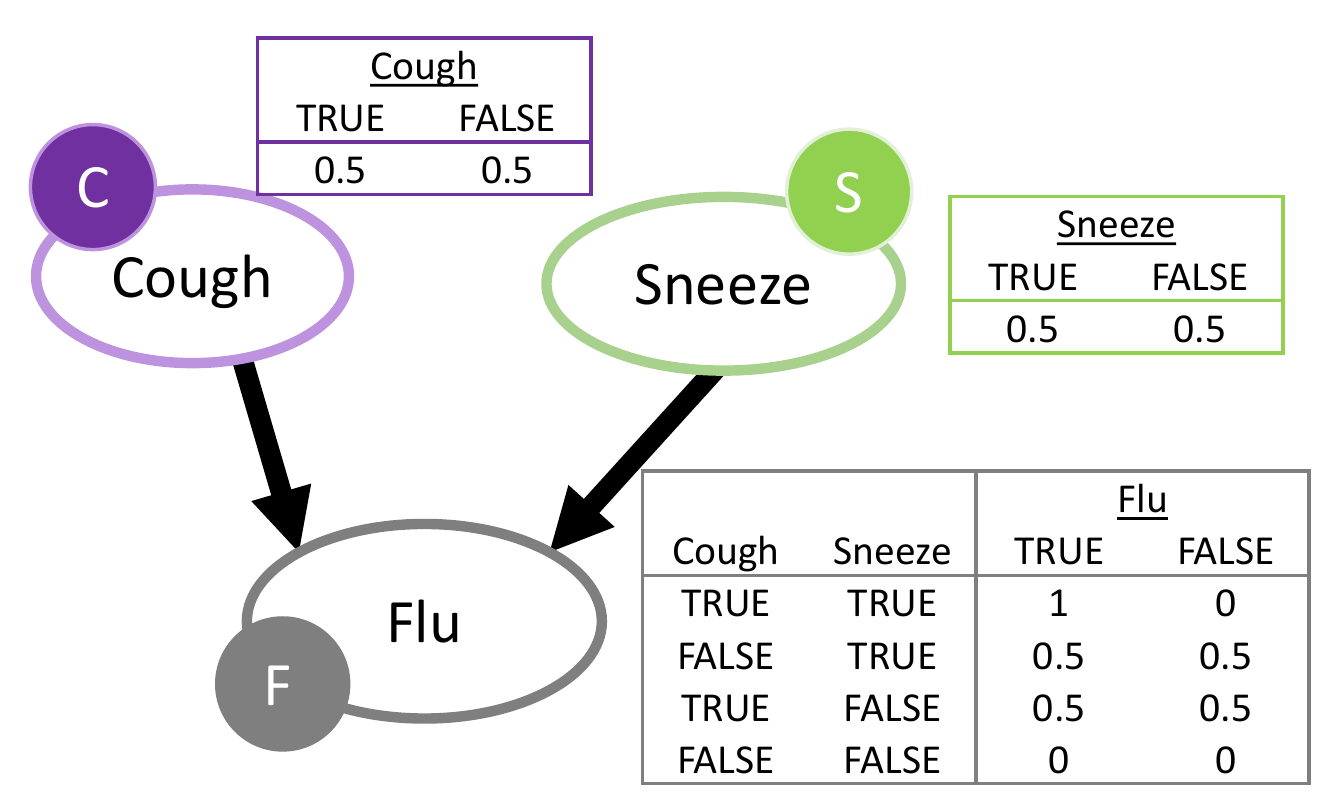}
	\end{center}
	\caption{Three node Bayesian network for diagnosing flu.  Symptoms of flu is given as coughing and sneezing.  Flu is diagnosed if both symptoms are detected.}
	\label{fig:flu}
\end{figure}

Using a simple three-node Bayesian network (Figure \ref{fig:flu}), we can compute the probability of flu using coughing and sneezing symptoms.

Using the confirmed symptoms of coughing and sneezing, the probability of a flu is  100\%, based on the result from the conditional probability table of the flu node.  However, given real-world situations, it may  be impossible to be 100\% accurate in detecting symptoms.  In this paper, we performed action classification which includes both the coughing and sneezing action.  Because the Reliability of classifying coughing and sneezing action is not perfect, the Bayesian inference must be penalized based on the system decision Reliability  and the risk of incorrect classification.  The base inference of the probability of diagnosing a flu is performed using the conditional probability table of the Flu node.
\begin{eqnarray}
\text{Pr(Flu)} &=& \frac{\text{Pr(Cough)} + \text{Pr(Sneeze)}}{2} \nonumber\\
	&=& \frac{0.783 +0.817}{2} =\fbox{0.800}\nonumber
\end{eqnarray}
Without risk, the probability of diagnosing a flu is \fbox{0.800}. By accounting for risk, we have to weaken this probability through multiplicative or additive means.  An example of multiplicative penalty due to risk is as follows:
\begin{eqnarray}
\text{Pr(Flu)} &=& \frac{\text{Pr(Cough)} + \text{Pr(Sneeze)}}{2} \cdot \frac{1}{1+\text{Risk}}\nonumber\\
&=& \frac{0.783 +0.817}{2} \cdot \frac{1}{1+\text{0.311}}=\fbox{0.610}\nonumber
\end{eqnarray}
Accounting for risk, the decision support system yields a \fbox{0.610} probability of detecting a flu.  {It can be seen that when $\text{Risk}=0$, the probability of diagnosing a flu is not influenced by the Risk of decision error.  This occurs when there is an absence of error from the machine learning model.  At this stage, errors derived from operators, users, or experts are not accounted for.}

\section{Summary and conclusion}\label{sec:conclusions}

Our study focuses on combining machine learning and machine reasoning techniques to 

 (a) improve the overall performance of recognizing various actions, 

 (b) determine the existence of biases and how to exploit these biases, and

 (c) diagnose the disease given evidence of selected symptoms.  

A modified residual temporal convolutional network guided by knowledge distillation wass proposed to perform action recognition, with accuracy of 82.5\%.  Using this architecture, we compare different testing protocols to determine the overall decision Reliability given the BIISC dataset.

The existence of bias within the given dataset is determined by examining the influence of various attributes on performance, specifically gender, pose, and view attributes.   Bias attributes can be exploited when testing information is known. Specifically for the video `viewing' attribute, an accuracy of 88.1\% was achieved on the `right' view cohort.  Given such bias in the system, the Risk of making an erroneous decision is seen as a function of the impact and cost of such decision.  

An application for diagnosing the flu symptoms is proposed using a combination of Bayesian network, machine reasoning, and machine learning.  The machine learning model provides evidence that is then evaluated for reliability using reasoning techniques.  The Bayesian network combines evidence with reliability to compute the final probability of flu.

\section*{Acknowledgments}
\begin{small}
This Project was partially supported by Natural Sciences and Engineering Research Council of Canada (NSERC) through grant ``Biometric-Enabled Identity Management and Risk Assessment for Smart Cities'', and the Department of National Defence’s Innovation for Defence Excellence and Security (IDEaS) program, Canada. 
\end{small}

{\small
	\bibliographystyle{IEEEtran}
	\bibliography{bias}
}

\end{document}